%% file: acl2023.tex
\pdfoutput=1

\documentclass[11pt]{article}

\usepackage[]{ACL2023}

\usepackage{times}
\usepackage{latexsym}
\usepackage{graphicx}
\usepackage{adjustbox}
\usepackage{booktabs}
\usepackage{multirow}
\usepackage{amsmath}
\usepackage{todonotes}
\usepackage[ruled,vlined]{algorithm2e}
\usepackage[T1]{fontenc}

\usepackage[utf8]{inputenc}

\usepackage{microtype}
\usepackage{amssymb}
\usepackage{pifont}

\usepackage[T1]{fontenc}

\usepackage[utf8]{inputenc}

\usepackage{microtype}

\usepackage{inconsolata}

\title{Is a Video worth $n\times n$ Images? A Highly Efficient Approach to Transformer-based Video Question Answering}

 \author{
  \textbf{Chenyang Lyu}$^\dag$
  ~~~~ \textbf{Tianbo Ji}$^\ddag$~~~~ \textbf{Yvette Graham}$^\P$~~~~ \textbf{Jennifer Foster}$^\dag$~~~~  \\
  $^\dag$ School of Computing, Dublin City University, Dublin, Ireland \\
  $^\ddag$ Nantong University, China \\
  $^\P$ School of Computer Science and Statistics, Trinity College Dublin, Dublin, Ireland\\
  \texttt{chenyang.lyu2@mail.dcu.ie}, \texttt{ygraham@tcd.ie}, \texttt{jennifer.foster@dcu.ie} \\
  \texttt{{jitianbo}@ntu.edu.cn}
}

\begin{document}
\maketitle

\input{0-Abstract.tex}
\input{1-Introduction.tex}
\input{2-Methodology.tex}
\input{3-Experiments.tex}
\input{4-Conclusion-and-Future-Work.tex}

\section*{Limitations}
The two VideoQA datasets used in experiments are associated relatively shorter videos, therefore it would be better if more experiments can be conducted on VideoQA datasets with long videos to verify the effectiveness of our approach on a wider range of VideoQA tasks. Although the proposed approach in this paper can also be used in other video-language tasks, our experiments focuses on a specific video-language task - VideoQA. Experiments on more video-language tasks are needed to show that our approach are also effective in other video-language tasks.

\bibliography{anthology,custom,trafficqa_arxiv}
\bibliographystyle{acl_natbib}

\input{5-Appendix.tex}
\end{document}

%% file: 0-Abstract.tex
\begin{abstract}
Conventional Transformer-based Video Question Answering~(VideoQA) approaches generally encode frames independently through one or more image encoders followed by interaction between frames and question. However, such schema would incur significant memory use and inevitably slow down the training and inference speed. In this work, we present a highly efficient approach for VideoQA based on existing vision-language pre-trained models where we concatenate video frames to a $n\times n$ matrix and then convert it to one \textit{image}. By doing so, we reduce the use of the image encoder from $n^{2}$ to $1$ while maintaining the temporal structure of the original video. Experimental results on MSRVTT and TrafficQA show that our proposed approach achieves state-of-the-art performance with  nearly $4\times$ faster speed and only 30\% memory use. We show that by integrating our approach into VideoQA systems we can achieve comparable, even superior, performance with a significant speed up for training and inference. We believe the proposed approach can facilitate VideoQA-related research by reducing the computational requirements for those who have limited access to budgets and resources. Our code will be made publicly available for research use.

\end{abstract}

%% file: 1-Introduction.tex
\section{Introduction}
 Transformer-based Video Question Answering~(VideoQA)~\cite{xu2016msrvtt,yu2018joint-msrvtt-mc,xu2021sutd-trafficqa, bain2021frozen,lei2022revealing} approaches relying on large scale vision transformers~\cite{dosovitskiy2020image-vit} have achieved strong performance in recent years. 
However, such approaches typically encode multiple video frames separately through one or more \textit{Image Encoders}~\cite{lei2021less-clipbert,luo2021clip4clip,xu2021videoclip,arnab2021vivit,zhong2022videoqa-survey} followed by interaction with question representations. This requires significant memory use and inevitably slows down training and inference speed. 
In order to reduce the computational cost required for modeling video representations from frames, we propose to arrange the frames sampled from one video as a single image. Specifically, we sample $n^{2}$ frames from one video and concatenate them as a single image with $n\times n$ grids. 

\begin{figure}
    \centering
    \includegraphics[width=\linewidth]{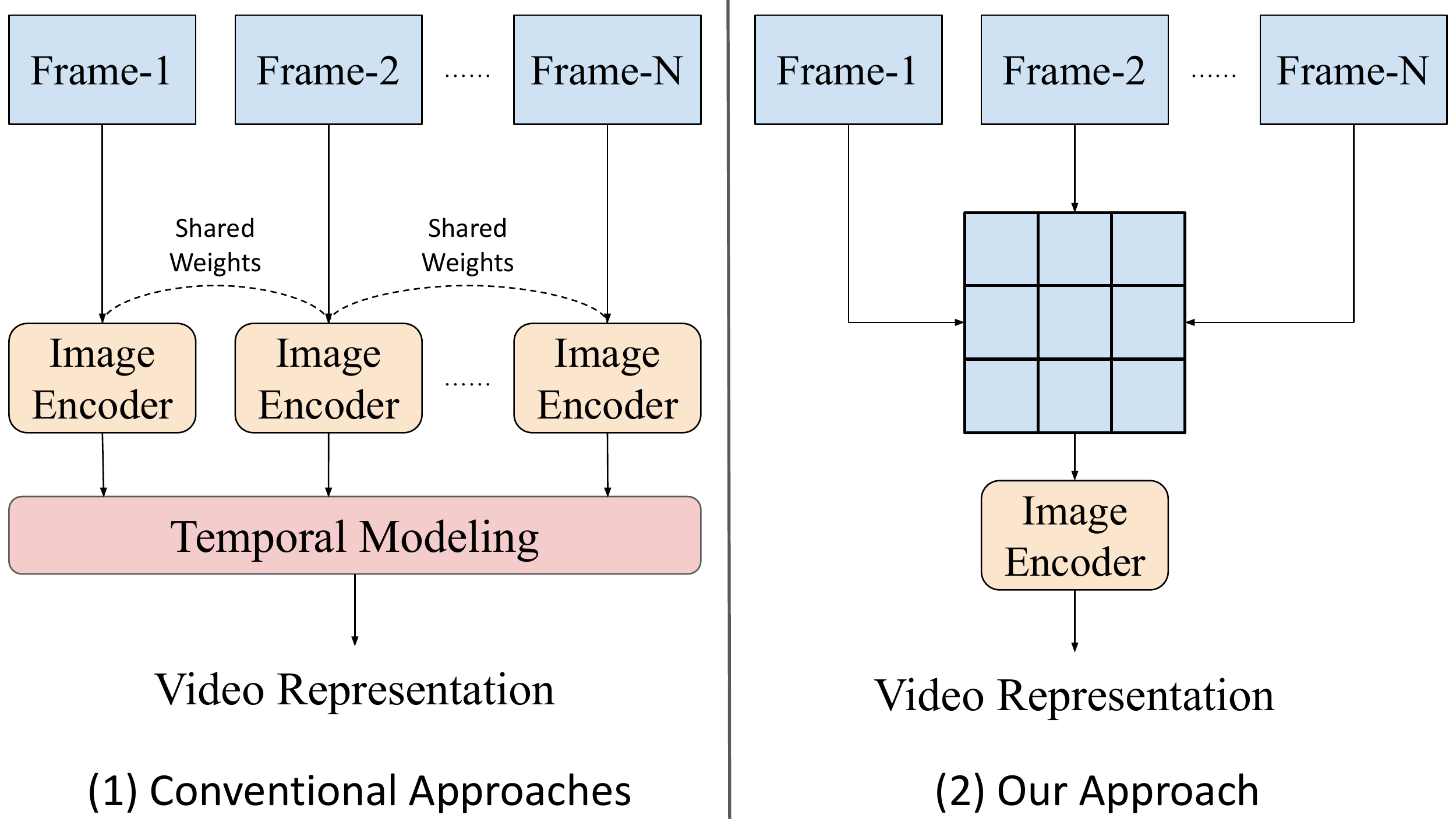}
    \caption{Conventional approach to encoding video frames for VideoQA and our proposed method. }
    \label{fig:introduction-figure}
\end{figure}

Figure \ref{fig:introduction-figure} shows the difference between our method (right) and conventional methods (left). In general, conventional approaches either independently encode video frames ~\cite{luo2021clip4clip,xu2021videoclip,lei2021less-clipbert,bain2022clip-hitchhiker} which require $N$ forward passes, or encode the sequence of all patches in video frames~\cite{bain2021frozen,arnab2021vivit} which quadratically increases the computational cost in the attention modelling. Both types of aforementioned encoding approaches can be expected to negatively impact training and inference speed, whereas our proposed method reduces this need substantially, now requiring only a single forward pass.

Diverging from previous approaches, our method has two distinctions: 1) it fully relies on existing available pre-trained vision-language models such as CLIP~\cite{radford2021learning-clip} without need for extra pre-training~\cite{bain2021frozen,lei2022revealing}; 2) it considers a multi-frame video as a single image, dispensing with the need for positional embedding at the frame level~\cite{bain2021frozen}, so only minor modifications to pre-trained models are necessary. 



More importantly, our approach has three advantages: 1) higher computational efficiency~\footnote{For example, the computational cost of~\newcite{bain2021frozen,arnab2021vivit} scales up quadratically w.r.t. the number of image patches whereas ours is invariant w.r.t. the number of image patches.}; 2) less memory use - our approach only uses an \textit{Image Encoder}  a single time; 3) our approach can be easily scaled up for large numbers of frames for long videos. Our approach also models a multiple-frame video as a single image while still (partially) maintaining the temporal structure of the original video. 

To validate the effectiveness of our approach, we conduct experiments on two benchmark VideoQA datasets: MSRVTT-MC~\cite{xu2016msrvtt,yu2018joint-msrvtt-mc} and TrafficQA~\cite{xu2021sutd-trafficqa}. Results show that our approach achieves comparable or even superior performance compared to existing models with nearly 4$\times$ faster training and inference speed and vast reduction in memory use~(30\%). Our contribution can be summarised as follows:
\begin{itemize}
    \item We propose a novel approach combining video frames as a single image to accelerate VideoQA systems ;
    \item Experimental results on MSRVTT-MC and TrafficQA show that our proposed approach achieves competitive performance with faster training-inference speed and lower memory use;

    
    \item We include additional experiments investigating options for arrangement of video frames for VideoQA;


    
\end{itemize}

%% file: 2-Methodology.tex
\section{Model Architecture}
\begin{figure}[t]
    \centering
    \includegraphics[width=\linewidth]{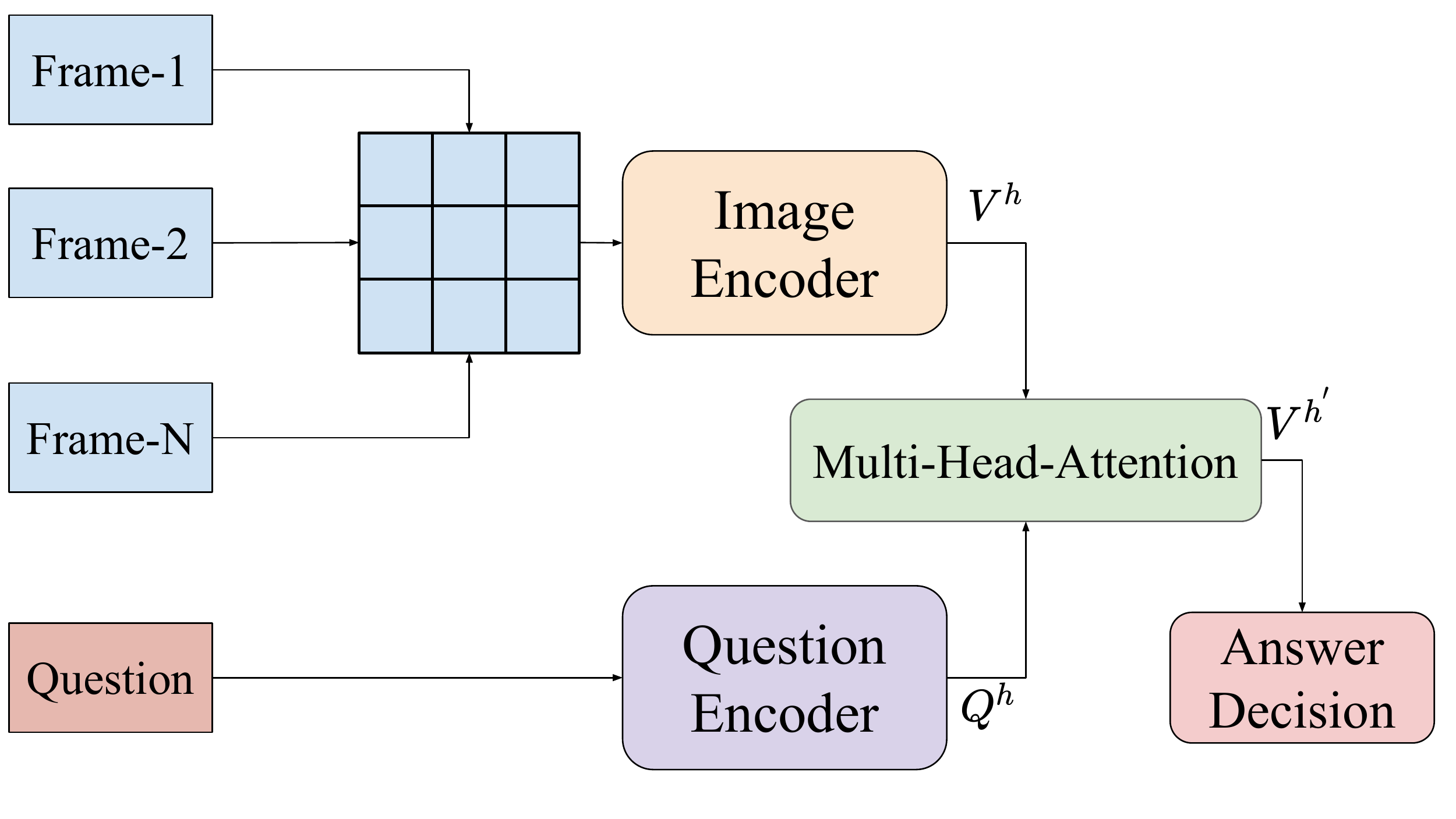}
    \caption{An overview of our proposed approach.}
    \label{fig:main_figure}
\end{figure}

In this section, we introduce details of our approach, of which an overview is shown in Figure~\ref{fig:main_figure}.

\subsection{Vision Transformer}
\label{sec:vision-transformer}
Generally Vision Transformer~(ViT)~\cite{dosovitskiy2020image-vit} flattens a single image to $m$ non-overlapping patches $v = \{p_{0}, p_{1},......,p_{m-1}\}$. All patches are fed into a linear projection and then regarded as discrete tokens in~\cite{transformer,gpt,bert} followed by transformer-based modeling~\cite{transformer,dosovitskiy2020image-vit}. The output feature is $v^{h} = \{h_{0}, h_{1},......,h_{m-1}\}$, where $E^{h}\in\mathbf{R}^{m\times d}$, $d$ is the dimension of the output feature for each patch.

For encoding a multiple-frame video, suppose that we have an input video $V=\{v_{0}, v_{1},......,v_{n-1}\}$ with $n$ frames, so for each frame $v_{i}$ ViT flattens it to $m$ non-overlapping patches $v_{i} = \{p_{i,0}, p_{i,1},......,p_{i,m-1}\}$. The patches for each frame are concatenated to form a sequence of patches:
\begin{multline}
        V=\{p_{0,0},......,p_{0,m-1},......,p_{i,0}, \\ p_{i,1},......,p_{i,m-1},......\}
\end{multline}

 which are fed into (a) linear projection(s) followed by transformer-based modeling~\cite{transformer,dosovitskiy2020image-vit}\footnote{Frames can be encoded separately through one or more \textit{Image Encoder}~\cite{luo2021clip4clip,xu2021videoclip,bain2022clip-hitchhiker} or all patches can also be concatenated and passed into one \textit{Image Encoder}~\cite{bain2021frozen,arnab2021vivit}}. We thus obtain frame-level representations:

\begin{equation}
    V^{h}=\{v_{0},v_{1},......,v_{n-1}\}
\end{equation}

where $V^{h}\in\mathbf{R}^{n\times d}$.~\footnote{Patch-level representations $V^{h}=\{h_{0,0},......,h_{0,m-1},......,h_{i,0}, h_{i,1},......,h_{i,m-1},......\}$ are used in~\cite{lei2021less-clipbert,bain2021frozen}}

\subsection{Interaction with Question Representations}

For a natural language question $Q=\{w_{0},w_{1},......,w_{k-1}\}$ consisting of $k$ words, we use a textual transformer to encode $Q$ to obtain a sentence-level representation $Q^{h}\in\mathbf{R}^{1\times d}$.
Since in this work, we mainly focus on reducing the computational cost of encoding videos, we perform simple interactions between video representations $V^{h}$ and question representations $Q^{h}$:

\begin{equation}
    V^{h^{'}} = \textsc{Multi-Head-Attention}(Q^{h},V^{h},V^{h})
\end{equation}
where $V^{h^{'}}\in\mathbf{R}^{1\times d}$ is the question weighted representations and \textsc{Multi-Head-Attention}~\cite{transformer} performs attention between $V^{h}$~(key and value) and $Q^{h}$~(query).     
\subsection{Frames Transformation} 

When encoding multiple frames, the encoding schema in~\ref{sec:vision-transformer} incurs significant memory use and additionally impedes training and inference speed~\cite{lei2021less-clipbert,luo2021clip4clip,xu2021videoclip,lei2022revealing}. Therefore, we propose a novel strategy to reduce the computational cost associated with encoding videos by combining all frames into a single image arranged by $n\times n$ grids. Practically, we arrange all frames to a matrix, $M$, in which each entry corresponds to a frame. For example, for a video with $n\times n$ frames, we put each frame into $M_{i,j}$ in a specific order. For example, frames can be arranged in $M$ via ascending or descending order~(either vertically or horizontally) based on its index in the video.~\footnote{Frames can be arranged via different order, more results are shown in Sec~\ref{sec:arrangement}} Next, we convert $M$ to a single image. Therefore, regardless of how many frames we use, the number of tokenized patches~(image tokens) are always a constant number. So that the resulting VideoQA system is computationally efficient. 

%% file: 3-Experiments.tex
\section{Experiments}
\subsection{Datasets}

We conduct experiments on two benchmark datasets for VideoQA: MSRVTT-MC~\cite{xu2016msrvtt,yu2018joint-msrvtt-mc} and TrafficQA~\cite{xu2021sutd-trafficqa}, which are multi-choice VideoQA datasets - each video in MSRVTT-MC is associated with 5 candidate options whereas TrafficQA provides 4 options for each question. We follow the standard data split for MSRVTT-MC~\cite{xu2016msrvtt,yu2018joint-msrvtt-mc}, where evaluation data have 2,990 videos. TrafficQA contains 62,535 QA pairs and 10,080 videos. We follow the standard split of TrafficQA: 56,460 QA pairs for training and 6,075 QA pairs for evaluation.

\subsection{Experimental Setup}
We use CLIP ViT-B/16~\cite{radford2021learning-clip}~\footnote{https://openai.com/blog/clip/} to initialize our \textsc{Image-Encoder} and \textsc{Text-Encoder}. We evenly sample 9 frames from the videos in MSRVTT-MC and TrafficQA for the main experiment. We train our model by 20 epochs with a learning rate of 1e-6. The training batch size is 16. We use a maximum gradient norm of 1. The optimizer we used is AdamW~\cite{adamw}, for which the $\epsilon$ is set to $1\times10^{-8}$.

\input{Tables/Table-0-Results-on-MSRVTT.tex}
\input{Tables/Table-1-Results-on-TrafficQA.tex}

\subsection{Evaluation Results}
We show the evaluation results on MSRVTT-MC~\cite{xu2016msrvtt,yu2018joint-msrvtt-mc} in Table~\ref{table:results_msrvtt}. Furthermore, we conduct experiments on TrafficQA~\cite{xu2021sutd-trafficqa} and the results are shown in Table~\ref{table:results_trafficqa}. We also present the results of separately encoding video frames~(\textsc{Multi-Frame}) as in Figure~\ref{fig:introduction-figure}~(left) and our approach that combines multiple video frames into a single image~(\textsc{Single-Frame}). For \textsc{Single-Frame}, the frames are arranged to a matrix via horizontally descending order. The evaluation results show that our approach \textsc{Single-Frame} achieves comparable and even improved performance relative to strong baselines including VideoCLIP~\cite{xu2021videoclip}, All-in-One~\cite{wang2022all-in-one}, Singularity~\cite{lei2022revealing} and CMCIR~\cite{liu2022cross-event-level-reasoning-trafficqa}. \textsc{Single-Frame} obtains a significant speed up~(approaching $\times 4$) compared to \textsc{Multi-Frame} approach while maintaining competitive performance. The memory use of \textsc{Single-Frame} is only 30\% to \textsc{Multi-Frame}, which are compared on Nvidia GTX 3090. The results on two benchmark datasets have shown the effectiveness of our approach for improving the computational efficiency while maintaining  accuracy of VideoQA systems.

\subsection{Effect of Number of Frames}
\begin{figure}
    \centering
    \includegraphics[width=\linewidth]{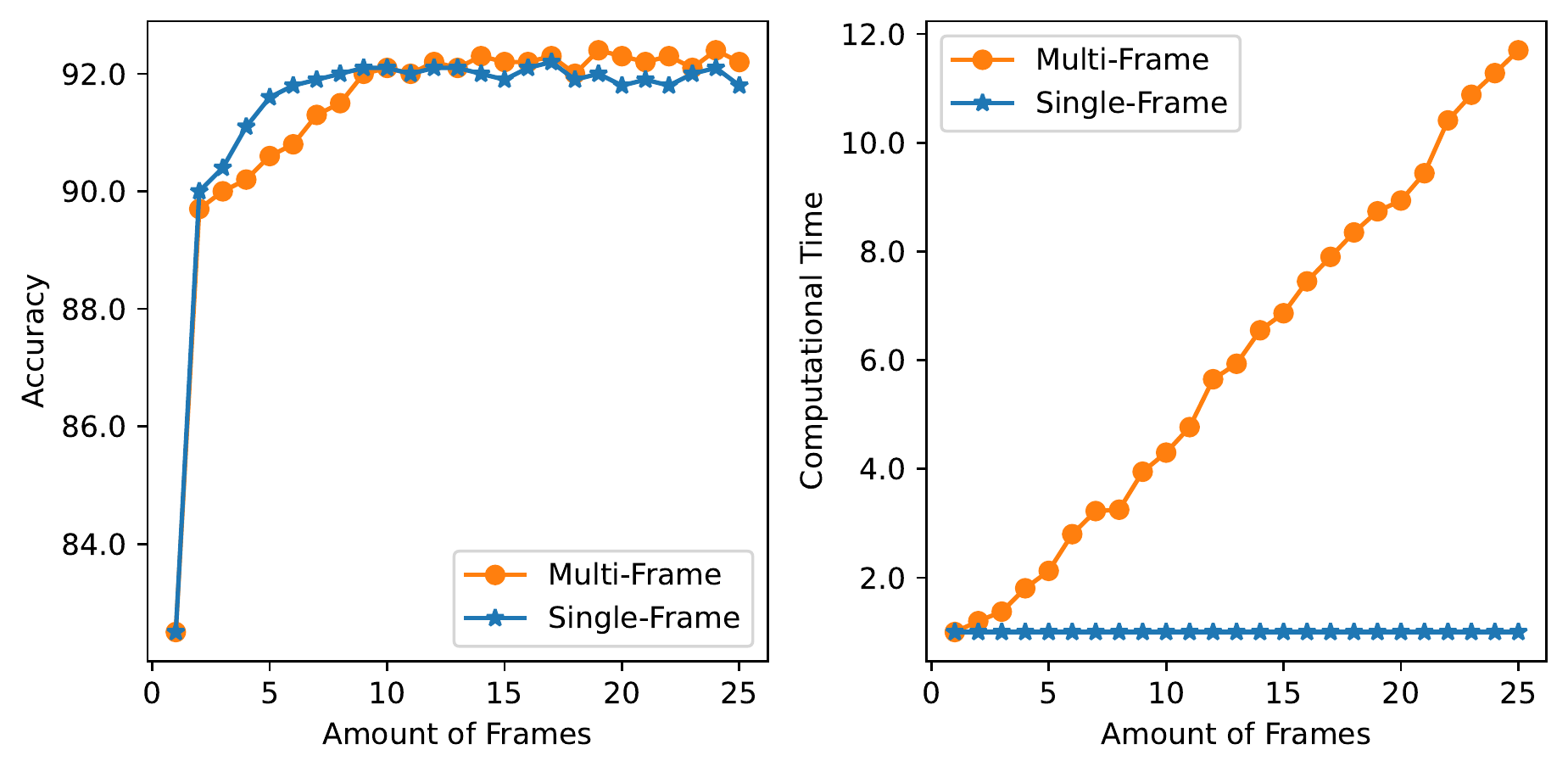}
    \caption{Evaluation results including accuracy~($\uparrow$) and computational time~($\downarrow$) on the effect of amount of
video frames on MSRVTT-MC.}
    \label{fig:effect_of_frames}
\end{figure}

We investigate the effect of the number of video frames used by our approach during the training and inference process, results are shown in Figure~\ref{fig:effect_of_frames}. We compare the performance of \textsc{Multi-Frame} and \textsc{Single-Frame} for number of frames ranging from 1 to 25~\footnote{For \textsc{Single-Frame} with a number of frames that is not a square number we up-sample it to the closest square number. For example, a \textsc{Single-Frame} that deals with 2 frames with index of $\{0, 1\}$, we upsample it to $\{0, 0, 1, 1\}$ as $2\times 2$ images.} in Figure~\ref{fig:effect_of_frames}. Results show that: 1) Both \textsc{Multi-Frame} and \textsc{Single-Frame} systems can benefit from more video frames; 2) \textsc{Single-Frame} is capable of achieving comparable and even better performance against \textsc{Multi-Frame}; 3) \textsc{Multi-Frame} costs much more computational time than \textsc{Single-Frame} especially when using a large number of video frames. Therefore, our proposed \textsc{Single-Frame} approach is able to achieve higher efficiency as well as competitive accuracy.

\subsection{Effect of Frame Order}
\label{sec:arrangement}
Furthermore, we investigate the effect of the arrangement of video frames used to form a single frame. The results of both the 9 frame and 16 frame configurations are shown in Table~\ref{table:results_frame_arrangement}. We report the results of \textsc{Vertical}, \textsc{Horizontal} and \textsc{Matrix} via either \textsc{ascending} or \textsc{descending} order.\footnote{Examples are shown in Appendix~\ref{appendix:frame_arrangement}.} The results in Table~\ref{table:results_frame_arrangement} reveal that both \textsc{Vertical} and \textsc{Horizontal} perform worst, and this is likely due to configurations distorting the visual information since they essentially squeeze the video frames either vertically or horizontally. The \textsc{Matrix} arrangement performs substantially better especially \textsc{Matrix(Horizontal-Descent)} and \textsc{Horizontal} generally yielding better performance compared to \textsc{Vertical} under \textsc{Matrix} arrangement.
\input{Tables/Table-3-Arrangement-of-Frames.tex}

%% file: Tables/Table-0-Results-on-MSRVTT.tex
\begin{table}
\resizebox{\linewidth}{!}{
\begin{tabular}{lcc}
\toprule
Models  & Accuracy   \\ \midrule

JSFusion~\cite{yu2018joint-msrvtt-mc}& 83.4 \\

ActBERT~\cite{zhu2020actbert}&  85.7 \\

ClipBERT~\cite{lei2021less-clipbert} & 88.2 \\

MERLOT~\cite{zellers2021merlot} & 90.9\\

VIOLET~\cite{fu2021violet}  & 90.9\\

VideoCLIP~\cite{xu2021videoclip} & 92.1\\

All-in-One~\cite{wang2022all-in-one} & 92.0\\

Singularity~\cite{lei2022revealing} & 92.1\\

Ours + \textsc{Multi-Frame} & 92.1~($1.0\times$) \\
Ours + \textsc{Single-Frame} & 92.2~($3.9\times$) \\
\bottomrule

\end{tabular}
}
\caption{Evaluation results on MSRVTT-MC~\cite{xu2016msrvtt,yu2018joint-msrvtt-mc} dataset. Number in bracket indicates the average of training and inference speed~($\uparrow$), which is evaluated on Nvidia GTX 3090.}
\label{table:results_msrvtt}
\end{table}











%% file: Tables/Table-1-Results-on-TrafficQA.tex
\begin{table}
\resizebox{0.95\linewidth}{!}{
\begin{tabular}{lccc}
\toprule
Models & Accuracy   \\ \midrule
Q-type (random)~\cite{xu2021sutd-trafficqa}& 25.0 \\ 
QE-LSTM~\cite{xu2021sutd-trafficqa} & 25.2   \\ 
QA-LSTM~\cite{xu2021sutd-trafficqa} & 26.7    \\ 
Avgpooling~\cite{xu2021sutd-trafficqa} & 30.5 \\ 
CNN+LSTM~\cite{xu2021sutd-trafficqa} & 30.8   \\ 
I3D+LSTM~\cite{xu2021sutd-trafficqa} & 33.2  \\

VIS+LSTM~\cite{ren2015exploring} & 29.9\\

BERT-VQA~\cite{Yang_2020_WACV} & 33.7 \\ 

TVQA~\cite{lei2018tvqa} & 35.2 \\

HCRN~\cite{Le_2020_CVPR} & 36.5 \\ 

Eclipse~\cite{xu2021sutd-trafficqa} & 37.0 \\ 

ERM~\cite{zhang2022erm-trafficcqa} & 37.1 \\

TMBC~\cite{luo2022temporal-trafficqa} & 37.2 \\

CMCIR~\cite{liu2022cross-event-level-reasoning-trafficqa} & 38.6\\

Ours + \textsc{Multi-Frame} & 39.7~($1.0\times$)  \\
Ours + \textsc{Single-Frame} & 39.7~($3.8\times$)  \\
\bottomrule
\end{tabular}
}
\caption{Evaluation results on SUTD-TrafficQA~\cite{xu2021sutd-trafficqa} dataset. Number in bracket indicates the average of training and inference speed~($\uparrow$).}
\label{table:results_trafficqa}
\end{table}


%% file: Tables/Table-3-Arrangement-of-Frames.tex
\begin{table}
\resizebox{0.98\linewidth}{!}{
\begin{tabular}{lccc}
\toprule
Arrangement of Frames & 9 frames & 16 frames  \\ \midrule

Vertical-Ascent & 89.8 & 89.9\\
Vertical-Descent & 89.1 & 89.5\\

Horizontal-Ascent & 88.7 & 88.8 \\
Horizontal-Descent & 87.5 & 88.6 \\

Matrix~(Vertical-Ascent) & 91.4 & 91.1\\

Matrix~(Vertical-Descent)   & 91.9 & 91.8\\

Matrix~(Horizontal-Ascent) & 91.6 & 91.2\\

Matrix~(Horizontal-Descent) & 92.2 & 92.1\\

Matrix~(Random)  & 90.5 & 90.7\\

\bottomrule

\end{tabular}
}
\caption{Evaluation results on the effect of the arrangement of video frames on MSRVTT-MC.}
\label{table:results_frame_arrangement}
\end{table}

%% file: 4-Conclusion-and-Future-Work.tex
\section{Conclusion and Future Work}
In this paper, we propose a highly efficient method for VideoQA where we combine multiple video frames into one single image. By adapting our approach, the computational cost of VideoQA systems can be significantly reduced. To validate the effectiveness of our approach, we conduct experiments on two benchmark datasets, MSRVTT-MC and TrafficQA. Results show that our approach achieves competitive performance and faster training and inference speed~(nearly $4\times$ faster) and less memory consumption~(30\%). Our approach provides a way  of significantly accelerating training and inference. In the future, we aim to explore how to adapt our approach to VideoQA with longer videos and additional video-related NLP tasks.

%% file: 5-Appendix.tex
\appendix

\section{Appendix}
\label{sec:appendix}

\subsection{Examples of frames with different arrangement order}
\label{appendix:frame_arrangement}
\begin{figure*}
    \centering
    \includegraphics[width=\textwidth]{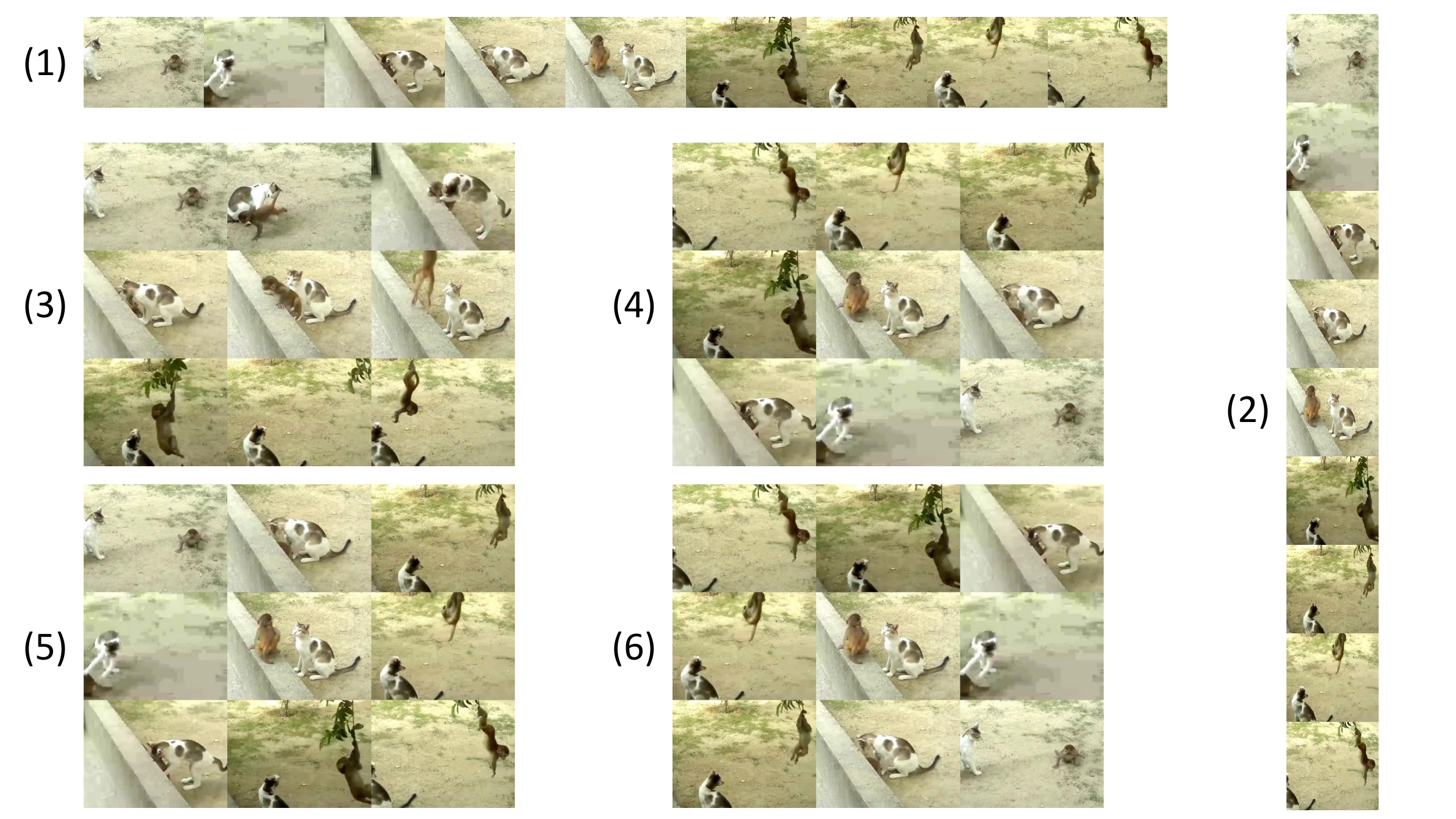}
    \caption{Frames arranged via different order.}
    \label{fig:frame_order_examples}
\end{figure*}

\begin{figure*}
    \centering
    \includegraphics[width=\textwidth]{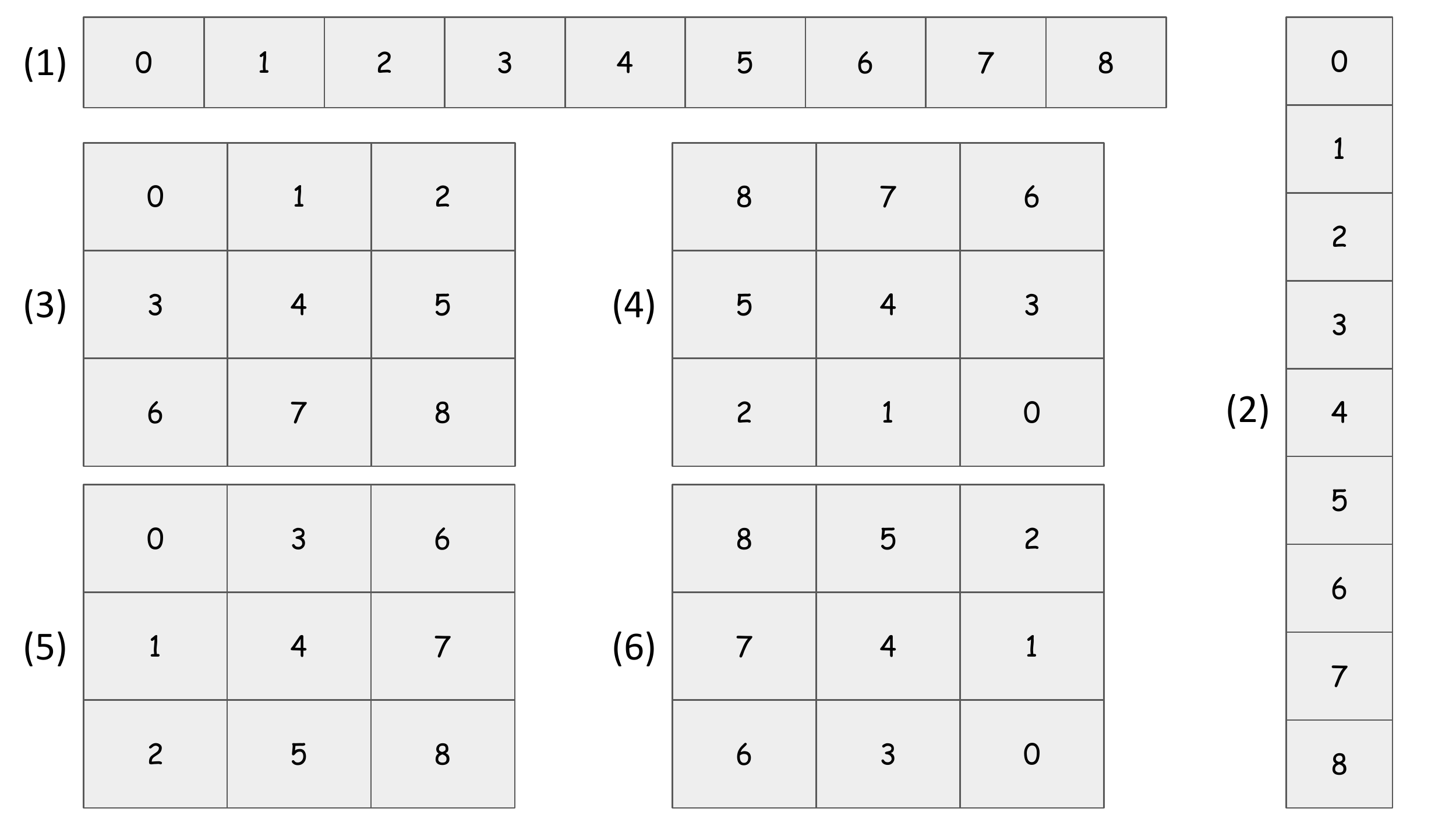}
    \caption{Corresponding frame index for different arrangement order.}
    \label{fig:frame_order_examples_index}
\end{figure*}

We present some examples of frames with different arrangement order in Figure~\ref{fig:frame_order_examples}, where we use 9 frames as examples. The arrangement orders are: (1) \textsc{Horizontal}. (2) \textsc{Vertical}. (3) \textsc{Matrix (Horizontal-Ascent)}. (4) \textsc{Matrix (Horizontal-Descent)}. (5) \textsc{Matrix (Vertical-Ascent)}. (6) \textsc{Matrix (Vertical-Descent)}. The corresponding video frame indices are shown in Figure~\ref{fig:frame_order_examples_index}.